# A utility-based spatial analysis of residential street-level conditions
## *A case study of Rotterdam*

Sander van Cranenburgh[1,*]
Francisco Garrido-Valenzuela[1]

[1]CityAI lab, Transport and Logistics Group,
Delft University of Technology
The Netherlands
*Corresponding author

*Abstract*
*Residential location choices are traditionally modelled using factors related to accessibility and socioeconomic environments, neglecting the importance of local street-level conditions. Arguably, this neglect is due to data practices. Today, however, street-level images –which are highly effective at encoding street-level conditions– are widely available. Additionally, recent advances in discrete choice models incorporating computer vision capabilities offer opportunities to integrate street-level conditions into residential location choice analysis. This study leverages these developments to investigate the spatial distribution of utility derived from street-level conditions in residential location choices on a city-wide scale. In our case study of Rotterdam, the Netherlands, we find that the utility derived from street-level conditions varies significantly on a highly localised scale, with conditions rapidly changing even within neighbourhoods. Our results also reveal that the high real-estate prices in the city centre cannot be attributed to attractive street-level conditions. Furthermore, whereas the city centre is characterised by relatively unattractive residential street-level conditions, neighbourhoods in the southern part of the city —often perceived as problematic— exhibit surprisingly appealing street-level environments. The methodological contribution of this paper is that it advances the discrete choice models incorporating computer vision capabilities by introducing a semantic regularisation layer to the model. Thereby, it adds explainability and eliminates the need for a separate pipeline to extract information from images, streamlining the analysis. As such, this paper's findings and methodological advancements pave the way for further studies to explore integrating street-level conditions in urban planning.*

## 1. Introduction

Residential location choices shape the infrastructure and functionality of cities. Specifically, these decisions individuals and households make about where to live have significant implications for transportation systems, housing markets, and urban planning (Cox & Hurtubia, 2021). Therefore, formulating policies to address urban challenges such as sprawl, housing affordability, and spatial inequality requires a thorough understanding of the factors influencing residential location choices (Pagliara & Wilson, 2010).

Residential location choices are commonly analysed using so-called Discrete Choice Models (DCMs) (Guevara & Ben-Akiva, 2006; Hunt, 2010; McFadden, 1977; Pérez et al., 2003). DCMs are grounded in random utility theory, where individuals are assumed to choose the alternative that maximises their utility from a set of discrete alternatives, each conceived as a bundle of attributes (e.g. cost, time, quality) (Ben-Akiva & Lerman, 1985). Researchers can estimate the parameters of the utility functions by confronting these models with empirical data, which involve trade-offs across the attributes. These parameters quantify the relative importance of each attribute in the decision-making. Residential location choices entail trade-offs along various factors, including (1) *Travel and accessibility-related factors*, such as the commute mode, commuting time, and distances to amenities like schools, stores, hospitals and playgrounds; (2) *Socioeconomic environments*, such as income levels, ethnic distribution, age and education level; and, (3) *Built environments and street-level conditions*, such as built density, housing types and typology, street layout, traffic safeness, greenness, parking conditions, and disorders (e.g. due to litter, graffiti, or weeds) (Giles-Corti et al., 2013).

Despite the acknowledged importance of the built environment and street-level conditions to residential location choices, they are commonly neglected in the analysis of residential location choices (Schirmer et al., 2014). This neglect relates to data practices. Residential location choice models have traditionally been built using census data, which does not contain detailed information about the built environment and street-level conditions. Only a limited number of residential location choice models have incorporated built environment variables, and those have typically relied on (tabular) cadastral data (Pinjari et al., 2007). While useful, these data often fall short of capturing detailed street-level conditions, such as the number of cars driving or parked in the street, the presence of sidewalks, or the amount and type of trees. Additionally, they lack information on more intangible street-level conditions, such as 'openness', 'scale, 'order', and 'continuity', all of which are known to influence how people perceive and evaluate streetscapes (Gjerde, 2011). As a result, comparatively little is known about how local street-level conditions – from a residential location choice perspective – are spatially distributed within cities.

Unlike tabular data, street-level images are particularly adept at encoding information about street-level conditions. The widespread use of street-level images on housing platforms and real-estate agency websites attests to the power of images to describe and convey information about street-level conditions as well as their importance to residential location choices. Importantly, nowadays, images containing information on street-level conditions are widely available from tech firms like Google, Apple, and Baidu's map services. Since these images became widely accessible, researchers have extensively utilised them to analyse and understand urban environments (Zhang et al., 2024). Notably, the pioneering work of Dubey et al. (2016) and subsequent studies have significantly advanced our understanding of how urban spaces are perceived in terms of e.g. safeness, vibrancy and liveliness using street-level images (Liu et al., 2017; Ma et al., 2021; Wei et al., 2022; Zhang et al., 2022; Zhang et al., 2018).

In addition to these developments, a new type of discrete choice model has recently been proposed, called Computer Vision enriched Discrete Choice Models (henceforth abbreviated as CV-DCMs) (Van Cranenburgh & Garrido-Valenzuela, 2023). CV-DCMs are capable of handling images as they integrate traditional random utility theory-based choice models and computer vision models. And, because the CV-DCM model proposed by Van Cranenburgh and Garrido-Valenzuela (2023) has been trained based on residential location choice data – which involved street-level images– it can be used to predict the utility derived from street-level conditions based on street-level images. Hence, it can be used to shed light on the spatial distribution of street-level conditions within a city.

The aims of this research are twofold: first, to shed light on the distribution of utility derived from street-level conditions in a residential location choice context at a city-level scale, and second, to examine the factors shaping this distribution. We use the city of Rotterdam as our case study. Rotterdam is the second largest city in the Netherlands, with about 670k inhabitants, and boasts a diverse array of neighbourhoods (Custers & Willems, 2024). To achieve our research aims, we capitalise on the widespread availability of street-level imagery and the recently developed CV-DCM. Specifically, we collect 300k geo-tagged street-level images of Rotterdam and calculate the utility derived from the street-level conditions using the CV-DCM. Finally, we aggregate the results at the postal code level to produce maps showing the spatial distribution.

This paper makes a substantive and methodological contribution. Substantively, it is the first to present a large-scale application of a CV-DCM, demonstrating its potential to generate new insights about *preferences* of urban environments. Thereby, it complements previous research on *perceptions* of the urban environment based on street-level imagery (Zhang et al., 2018). Although perceptions and preferences are closely related, they are distinct concepts. Preferences are grounded in the theory of choice behaviour (Lancaster, 1966; Luce, 1959; Samuelson, 1948) and govern what people choose (typically leading to economic demand). Perceptions are people's subjective interpretations of sensory stimuli, which can, but do not necessarily, influence individuals' choices. As a methodological contribution, we extend CV-DCMs to deliver insights into the factors underlying the utility distribution derived from street-level conditions. Specifically, we have added a semantic regularisation layer to the model. This layer is designed to predict key semantic attributes that are believed to influence residential location choices, such as the number of cars and the amount of visible sky, alongside the location decisions themselves. Thus, the extended model extracts the semantic attributes from images, which it, in turn, uses to explain the residential location choices. This integrated approach eliminates the need for a separate pipeline to process, segment, or otherwise extract information from images, thereby streamlining the analysis. Moreover, because the extended model is computationally efficient to deploy, it enables the analysis of a large number of images, making it feasible to assess the utility distribution derived from street-level conditions on a city-wide scale.

The remaining part of this paper is organised as follows. Section 2 describes the method, including the data collection for our case study. Section 3 presents the results and provides discussions in three sub-sections. Section 3.1 reports on the joint distributions of the semantic attributes. Section 3.2 presents the spatial distribution of utility derived from street-level conditions to residential location choices. Finally, section 3.3 zooms into areas of interest to investigate the contributing factors to its predictions for these specific areas, such as the number of cars or the amount of trees. The paper ends with a conclusion and a discussion of the limitations and avenues for future research.

## 2. Method

Figure 1 shows an overview of our methodology, which comprises three main steps:
1. Street-level image data retrieval.
   We collect URLs of Google Street View from the city of Rotterdam.

2. Applying a (trained) semantic CV-DCM to the collected street-level images.
   For the purpose of this paper, we extend the recently proposed CV-DCM model and retrain it.

3. Aggregating the results at a spatial unit of analysis.
   In this step, we aggregate the semantic CV-DCM's predictions in order to obtain information in a meaningful spatial unit.

In the section below, we discuss each step in more detail.

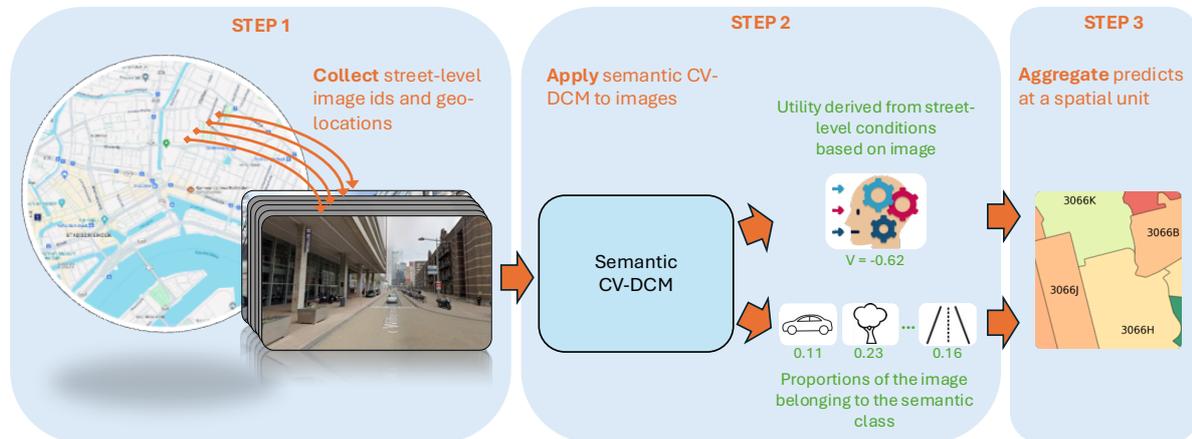

*Figure 1: Overview of methodology*

## 2.1. Image data retrieval for Rotterdam

Step 1 of the method involves the collection of URLs to street-level images of residential areas in the city of Rotterdam. To compile this dataset, we took the following steps:

1. We created a grid of points with 100-metre spacing within areas designated as residential areas within Rotterdam.
2. We retrieved street-level 360-degree panorama images of Rotterdam taken in the last decade (i.e. year ≥ 2014) following the process described in Van Cranenburgh and Garrido-Valenzuela (2023).
3. From each panorama, we generated two image URLs with 90-degree angles to the direction of the street (to both directions). This latter ensures that the images are 'window views', in line with the orientation of the images used by Van Cranenburgh and Garrido-Valenzuela (2023) to train their CV-DCM.
4. The municipality of Rotterdam has various suburbs and a large stretch of land area primarily designated for port activities. We excluded images from suburbs outside the city's core and neighbourhoods with primarily port activities. Additionally, we dropped images taken on highways and primary roads.
5. The final database contains of URLs from ~300k street-level images of residential streets in Rotterdam.

## 2.2. Semantic CV-DCM

Step 2 of our methodology involves applying a trained CV-DCM to the images collected in Step 1. To this end, this subsection introduces the CV-DCM (section 2.2.1), extends it (section 2.2.2), and reports on the training data (section 2.2.3) and training results (section 2.2.4) of the model that we apply.

### 2.2.1. CV-DCM

The CV-DCM assumes decision-makers, denoted by subscript *n*, make decisions based on Random Utility Maximising (RUM) principles (McFadden, 1974). Equation 1 shows the utility function of this model in its most general form. In this model, the utility of alternative *i*, denoted $U_i$, is derived from the numeric attributes $X_i$ and attributes encoded in the image $\mathcal{I}_i$, which is presented as part of the alternative, see Figure 2. In Equation 1, $f(\circ)$ is a function that maps the numeric attributes onto utility; $g(\circ)$ is a function that maps the attributes encoded in the image onto utility.

In line with the vast majority of discrete choice models, we assume utility is linear and additive, see Equation 2. The main advantage of these assumptions is that the $\beta_m$ parameters can be interpreted as marginal utilities; representing the *ceteris paribus* change in utility associated with a one-unit change in the corresponding explanatory variable $x_m$. The mapping $g(\circ)$ can be further decomposed into two functions: first, $\varphi(\circ)$ produces a feature map (aka embedding) of the image, denoted $Z_{in} = \{z_{i1n}, z_{i2n}, \ldots, z_{iKn}\}$. This mapping is carried out by a computer vision architecture, such as a Convolutional Neural Network (CNN) or a Vision Transformer (ViT). After that, the feature map is mapped linearly onto utility: $\sum_k \beta_k z_{ikn}$. Finally, the error term, $\varepsilon_{in}$, is assumed to be independent and identically Extreme Value Type I distributed with a variance of $\pi^2/6$. As a result, the choice probabilities take the well-known and convenient closed-form logit formula (aka Softmax), which enables efficient computation of the choice probabilties (Equation 3).

$$U_{in}(X_{in}, \mathcal{I}_{in}) = \overbrace{f(X_{in})}^{\substack{\text{utility derived from} \\ \text{numeric attributes}}} + \overbrace{g(\mathcal{I}_{in})}^{\substack{\text{utility derived from attributes} \\ \text{encoded in image}}} + \varepsilon_{in} \qquad \text{Equation 1}$$

$$U_{in}(X_{in}, \mathcal{I}_{in}) = \underbrace{\overbrace{\sum_m \beta_m x_{inm}}^{\substack{\text{utility derived from} \\ \text{numeric attributes}}} + \overbrace{\sum_k \beta_k z_{ikn}}^{\substack{\text{utility derived from attributes} \\ \text{encoded in image}}}}_{V_{in}} + \varepsilon_{in} \qquad \text{Equation 2}$$

where
$Z_{in} = \varphi(\mathcal{I}_{in})$
$\varepsilon_{in} = \sim EV\ type\ I$

$$P_{in} = \frac{e^{V_{in}}}{\sum_{j \in J} e^{V_{jn}}} \qquad \text{Equation 3}$$

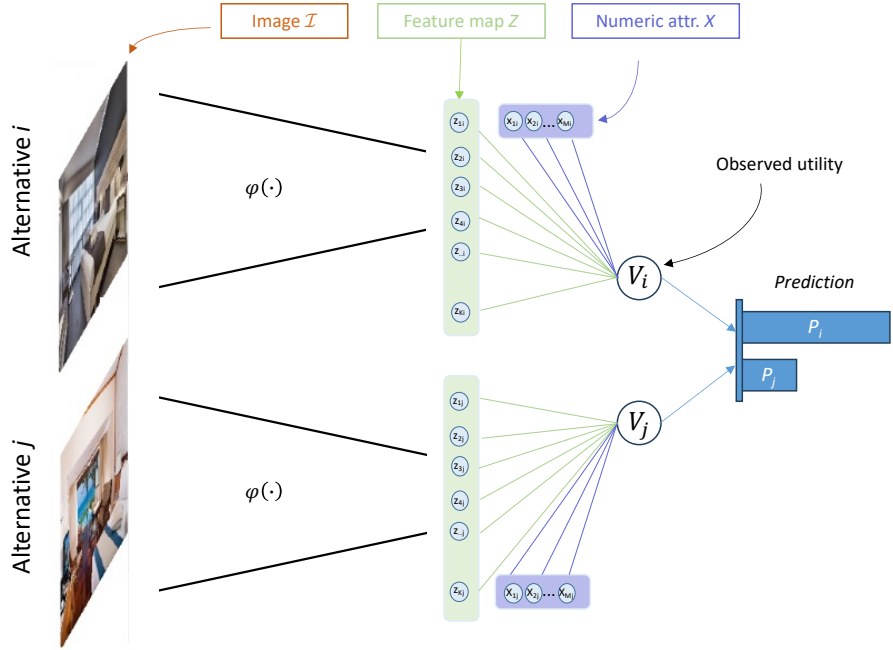

*Figure 2: Model structure of CV-DCM (for J =2).
Subscript n dropped for legibility.*

A drawback of the CV-DCM is that its computer vision part $g(\mathcal{I}_{in})$ is opaque. Even though the feature map is mapped linearly onto utility (see Equation 2), the $\beta_k$ parameters do not carry a behavioural meaning. This lack of interpretation is because the elements of the feature map, $z_{ikn}$, themselves do not carry a priori semantic meaning. Therefore, the CV-DCM does not allow tracing back which attributes encoded in a street-level image contribute to utility and to what extent. Put differently, the CV-DCM (of Figure 2) does not reveal the relative importance of, e.g., a tree or a parked car to the residential location choice. This opacity limits its usefulness for the purpose of this paper as it hampers the policy relevance of the model's outcomes for urban and transport planners involved in projects to (re)design the urban space

One way to obtain more behavioural insights about people's preferences over the attributes encoded in street-level images (i.e. image feature map) is to extract semantic objects, like cars and trees, from the image and feed these directly into the utility function of a traditional choice model. Such an approach is taken in adjacent studies, like Nagata et al. (2020); Ramírez et al. (2021). This approach allows for estimating the effect of the attributes on the utility function, offering a clear understanding of how specific objects in the environment influence choices. Such a top-down approach, however, is constrained by the analyst's predefined selection of objects, limiting the ability to detect unexpected or emergent elements. Furthermore, it cannot capture more nuanced visual cues, such as Gestalt principles—proximity, similarity, common fate, common region, closure, continuity, connectedness, and common orientation. These factors play a significant role in shaping how people perceive and interact with their surroundings and are widely applied in urban planning (Gjerde, 2011). For instance, Gestalt principles are often employed to organise buildings, streets, and public spaces in ways that appear visually balanced and cohesive. They are also used to design public areas and parks that feel well-defined, leveraging principles like proximity and closure to create inviting, functional environments.

### 2.2.2. Semantic CV-DCM

In light of the considerations in section 2.2.1, we propose to extend the CV-DCM to provide more behaviour insights while keeping the model capable of capturing nuanced visual cues. Specifically, we propose adding a semantic regularisation layer to the model that aims to predict semantic attributes that are believed to be important to the residential location choice, such as the cars, sky and trees, see Figure 3. The semantic attributes (nodes) are predicted from the feature map. This is possible because the feature map can be understood as a compact data representation of the image. In fact, in the field of representation learning, feature maps are intentionally created to be used for various downstream tasks, as they encapsulate a rich and diverse range of information. The semantic attributes, in turn, are assumed to map linear onto utility. Because the regularised nodes carry semantic meaning and are mapped linearly onto utility, the associated $\beta^{sem}$ parameters can be interpreted as the marginal utilities of the semantic objects. Thus, by adding the semantic regularisation layer, the model captures the utility of the semantic attributes in an explicit and transparent way (as opposed to via $\beta_k$).

Equation 4 formally introduces the Semantic CV-DCM. In this model, the systematic part of the utility, $V_{in}$, comprises three terms. The first term captures the utility derived from the numeric attributes in a traditional linear-additive fashion, with $\beta^{num}$ being the marginal utilities associated with the numeric attributes. The second term captures the utility derived from the semantic attributes, $S_{in} = \{s_{i1n}, s_{i2n}, \dots, s_{iTn}\}$, in a linear-additive fashion, with $\beta^{sem}$ being the marginal utilities associated with the semantic attributes. The third term captures the residual utility. That is, the utility derived from attributes encoded in the image which are not captured by the semantic term, including subtle visual cues such as gestalt. In contrast to $\beta^{num}$ and $\beta^{sem}$, the marginal utilities associated with the residual part of utility are not behaviourally interpretable. The semantic and residual utilities are computed from the feature map, $Z_i$, which is obtained through the mapping function $\varphi(\mathcal{I}_i|w_\varphi): \mathbb{R}^{H \times W \times C} \to \mathbb{R}^K$. The function $h(\circ): \mathbb{R}^K \to \mathbb{R}^T$ maps the feature map to the semantic attributes $S_{in}$.

$$U_{in} = \underbrace{\underbrace{\sum_m \beta_m^{num} x_{imn}}_{\text{Utility derived from numeric attributes}} + \underbrace{\sum_t \beta_t^{sem} s_{itn}}_{\text{Utility derived from semantic attributes encoded in the image}} + \underbrace{\sum_k \beta_k^{res} z_{ikn}}_{\text{Residual utility derived from attributes encoded in the image}}}_{V_{in}} + \varepsilon_{in} \qquad \text{Equation 4}$$

$$\text{where } Z_{in} = \varphi(\mathcal{I}_{in}|w_\varphi)$$

$$S_{in} = h(Z_{in}|w_h)$$

To be able to identify the marginal utilities associated with semantic attributes that are measured in proportions of the image, we need to consider the normalisation. Because all images have a fixed amount of pixels, information redundancy occurs if we would not apply a normalisation. To identify the model, we thus must fix one of the marginal utilities of $\beta^{sem}$ to zero. When we do so, all marginal utilities of semantic attributes (that are measured in proportions) must be interpreted relative to the marginal utility associated of the fixed class. In addition, it is important to note that $\beta^{sem}$ includes a parameter for the unsegmented proportion of images, $\beta_{unsegmented}^{sem}$. After all, not every pixel in each image is necessarily assigned to a semantic class, leaving an unsegmented proportion. For instance, street-level images may contain container boxes, construction sites, or just blurs, which may not be in the list of semantic attributes. The parameter, $\beta^{sem}_{unsegmented}$, is interpreted as the mean marginal utility of the unsegmented pixels relative to fixed class.

The proposed Semantic CV-DCM is trained to simultaneously predict the residential location choices from the Stated Choice (SC) experiment and the semantic attributes. Hence, the loss function of the Semantic CV-DCM comprises a loss associated with the residential location choices (specifically, the cross-entropy of the observed choices) and a loss associated with the predicted semantic attributes (specifically a RMSE loss), see equation 5. Importantly, in this equation, κ controls the balance between the two losses during the training. That is, setting κ to 0 implies the training focuses only on predicting the residential location choices; setting κ to 1 implies the training focuses only on predicting the semantic attributes.

$$Loss = (1-\kappa)\overbrace{\frac{-1}{N}\sum_{n=1}^{N}\sum_{j=1}^{J} y_{nj} \log(P_{nj}|X_{nj}, \mathcal{I}_{nj}, \beta)}^{cross\ entropy\ loss}$$

$$+ \kappa \sqrt{\overbrace{\frac{1}{N \cdot J \cdot S}\sum_{n=1}^{N}\sum_{j=1}^{J}\sum_{s=1}^{S}(Y_{njs} - \hat{Y}_{njs})^2}^{RMSE}}$$

Equation 5

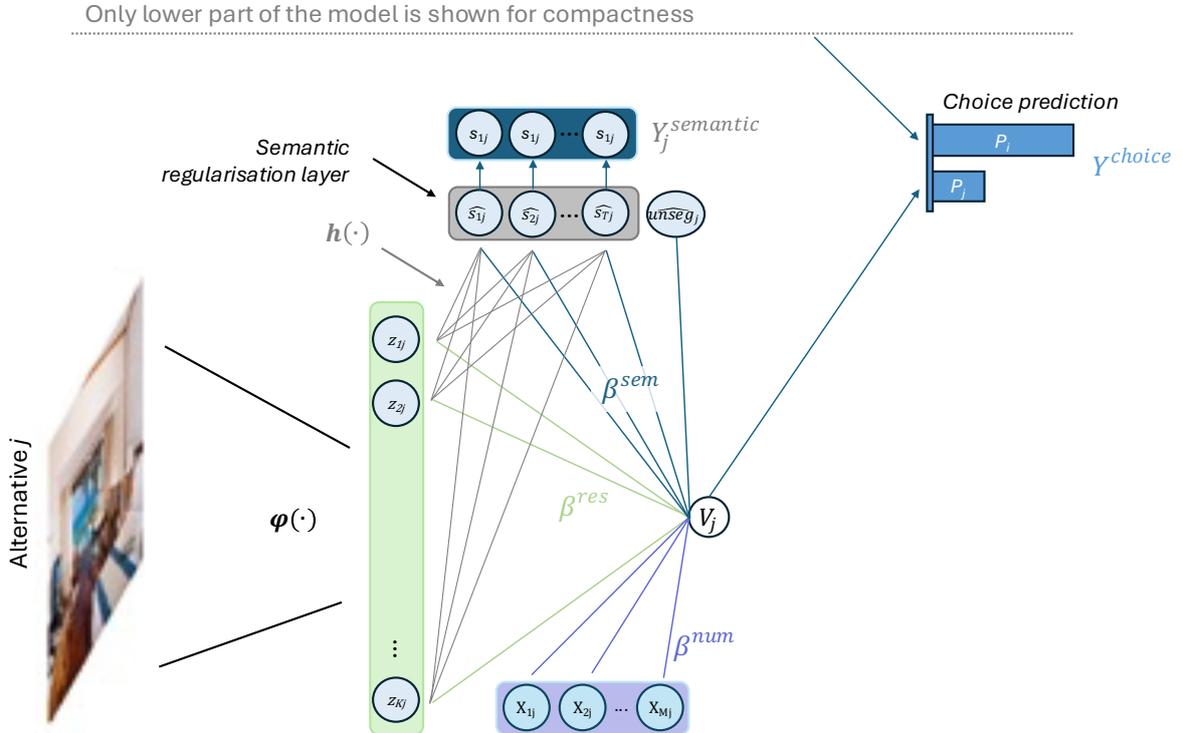

*Figure 3: Semantic CV-DCM (for J =2). Subscript n dropped for legibility.*

### 2.2.3. Training data

To train the Semantic CV-DCM, we need two sources of data: (1) a data set containing the choices over choice tasks with numeric and visually encoded attributes, and (2) a data set containing ground truth levels of the semantic attributes encoded in the images used in the choice data set. Regarding the former, in this study, we use the residential location choice data collected by Van Cranenburgh and Garrido-Valenzuela (2023), which have been made openly available at the paper's accompanying repository.[1] Figure 4 shows a screenshot of a choice task from their stated choice experiment. In this experiment, respondents had to make trade-offs between street-level conditions, presented using an image, and two numeric attributes: monthly housing cost and commute travel time. The street-level images were randomly drawn from a large database of street-level images. In the stated choice experiment, each participant was presented with 15 choice tasks. The stated choice experiment was conducted by a largely representative sample ($N = 800$) of the Dutch 18+ year old population. After cleaning, the data set contains 11,732 choice observations in which 7,341 unique street-level images are used. Furthermore, the data set has been split into a training and test set so that the images do not overlap between the two sets, ensuring that images in the training set are not present in the test set and vice versa.

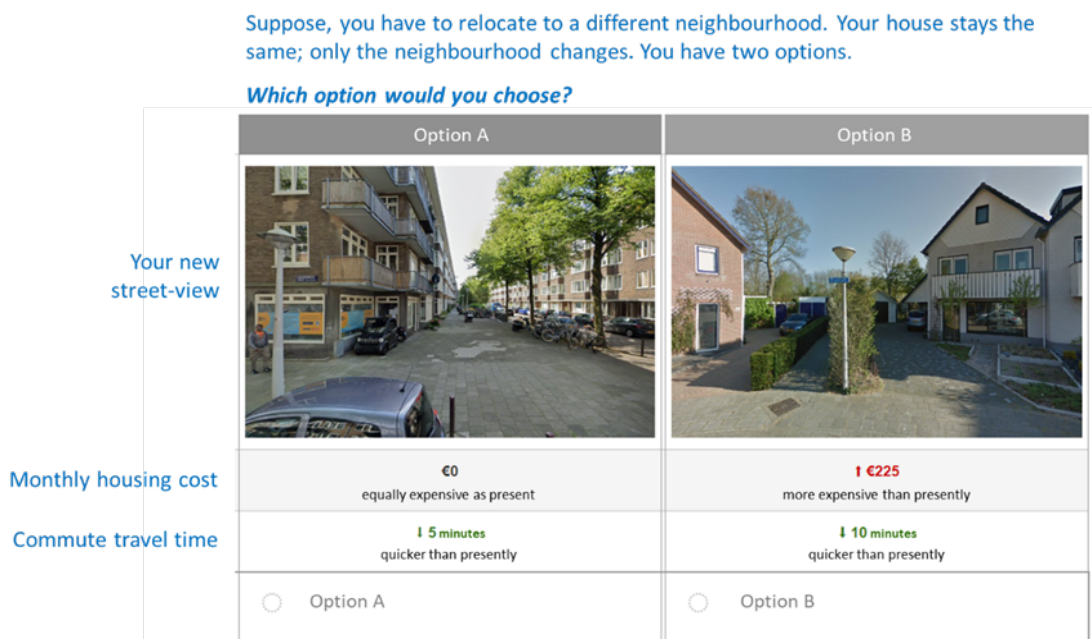

*Figure 4: Screenshot of the stated choice experiment (image source: Google) (translated to English)*

Regarding the data set containing the ground truth levels of the semantic attributes encoded in the images, we used state-of-the-art zero-shot object detection and segmentation models to construct it. Unlike traditional object detection models that have been trained using supervised learning, zero-shot models can detect objects beyond those predefined classes for which the models have been trained. More specifically, our detection and segmentation pipeline combines an object detection model called GroundingDINO and a segmentation model called Segment Anything Model (SAM). GroundingDINO is a zero-shot model for detecting objects developed by Liu et al. (2023). It integrates a transformer-based detection model called DINO (Zhang et al., 2022) and grounded pre-training. DINO achieves object detection based on text prompts. Grounded pre-training is crucial in enabling the model to acquire knowledge from textual descriptions of objects and scenes in a natural language context. The Segment Anything Model has been developed by Meta AI Research (Kirillov et al., 2023) with the aim of building a foundation model for image segmentation. While using text prompts as input is explored in the seminal paper, Meta AI has not released this capability. Thus, the XAI pipeline uses GroundingDINO (Liu et al., 2023) to identify objects based on a given text prompt and, subsequently, uses SAM to segment the identified objects and extract information regarding their quantity and number of pixels.

---

[1] https://github.com/TUD-CityAI-Lab/Computer-vision-enriched-DCMs

We applied this pipeline to all images that were used in the stated choice experiment, using the semantic attributes listed in Table 1. The selection of these attributes is informed by a brief literature review with the aim of finding semantic attributes which can be expected to be important to residential location choices. Due to the limited research specifically on residential location choices that account for street-level conditions, much of the relevant literature comes from the fast-growing body of literature about using street-level images to understand perceptions and behaviour (He & Li, 2021; Zhang et al., 2024). Numerous such studies associate street-level features with behaviours (e.g. walking, running, or criminal behaviour) or with perceptions (e.g. safety, liveliness, or beauty) (Dubey et al., 2016). Furthermore, we opted for a relatively concise set of semantic attributes, as initial trials indicated that expanding the set with additional, closely related attributes compromised the accuracy and quality of the segmentations. Figure 5 illustrates four examples of the segmented images based on this list of segments. Finally, only for the semantic attribute 'car' we use –besides the proportions– also the count. The rationale for this is that the count is meaningful for cars, as people may associate it with traffic safety, noise pollution, or air pollution, but not for the other semantic attributes.

*Table 1: List of semantic attributes*

| Semantic attributes (prompts) | Context of study | References |
|---|---|---|
| Car | Perception<br>Perception<br>Perception<br>Crimes rates | (Rossetti et al., 2019)<br>(Ramírez et al., 2021)<br>(Zhang et al., 2018)<br>(Zhanjun et al., 2022) |
| Building | Perception<br>Perception<br>Perception<br>Urban green space<br>Running<br>Drug places | (Rossetti et al., 2019)<br>(Ramírez et al., 2021)<br>(Zhang et al., 2018)<br>(Chen et al., 2024)<br>(Jiang et al., 2022)<br>(Zhou et al., 2021) |
| Grass | Perception<br>Urban Green space<br>Running<br>Walking | (Zhang et al., 2018)<br>(Chen et al., 2024)<br>(Jiang et al., 2022)<br>(Koo et al., 2022) |
| Road | Perception<br>Perception<br>Perception<br>Urban green space<br>Running<br>Crime rates | (Rossetti et al., 2019)<br>(Ramírez et al., 2021)<br>(Zhang et al., 2018)<br>(Chen et al., 2024)<br>(Jiang et al., 2022)<br>(Zhanjun et al., 2022) |
| Sky | Perception<br>Perception<br>Perception<br>Urban green space | (Rossetti et al., 2019)<br>(Ramírez et al., 2021)<br>(Zhang et al., 2018)<br>(Chen et al., 2024) |
| Trees | Perception<br>Perception<br>Urban green space<br>Running<br>Walking<br>Crime rates | (Ramírez et al., 2021)<br>(Zhang et al., 2018)<br>(Chen et al., 2024)<br>(Jiang et al., 2022)<br>(Koo et al., 2022)<br>(Zhanjun et al., 2022) |
| Plants | Running<br>Walking | (Jiang et al., 2022)<br>(Koo et al., 2022) |
| Fence | Perception<br>Perception<br>Urban green space<br>Running<br>Crimes rates | (Rossetti et al., 2019)<br>(Ramírez et al., 2021)<br>(Chen et al., 2024)<br>(Jiang et al., 2022)<br>(Zhanjun et al., 2022) |
| Water | Urban green space<br>Running<br>$NO_2$ prediction | (Chen et al., 2024)<br>(Jiang et al., 2022)<br>(Qi et al., 2022) |

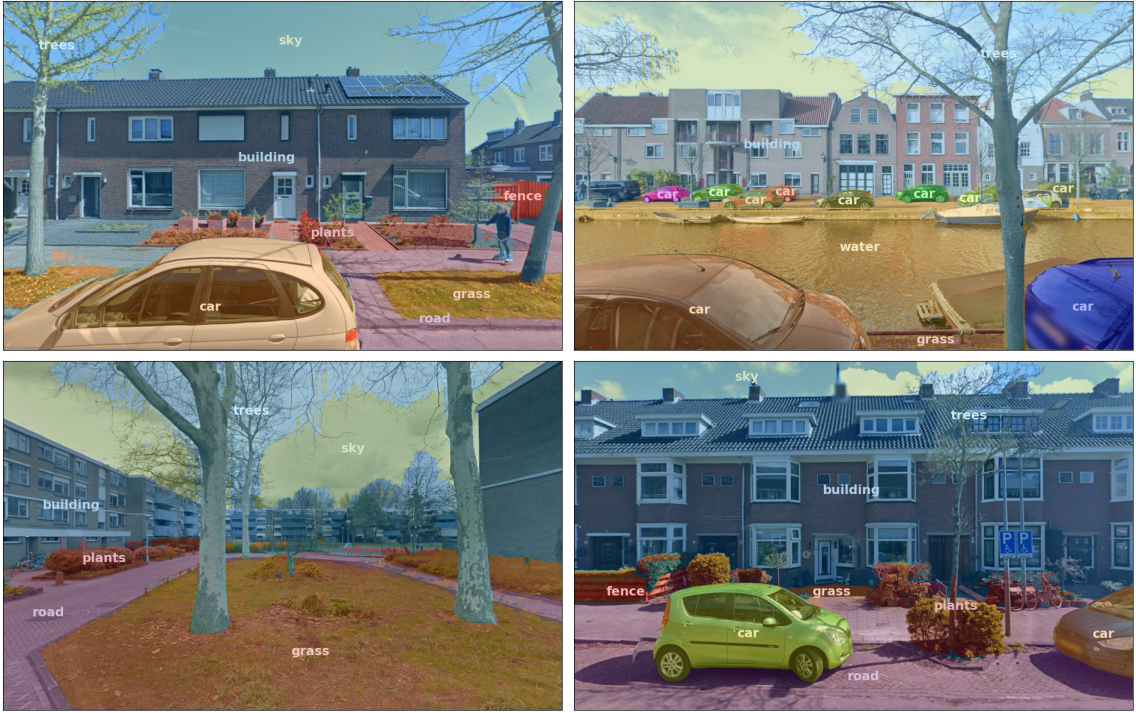

*Figure 5: Four examples of segmented images using GroundingDINO and SAM*

### 2.2.4. Training results

Training the Semantic CV-DCM is a sequential process. The reason that we must train this model sequentially, as opposed to jointly, relates to isomorphic information sharing (Sifringer & Alahi, 2023). Isomorphic information sharing may occur because the model can explain the utility associated with a semantic attribute, say car, directly via $\beta^{res}$ or 'indirectly' via $\beta^{sem}$ (which is what we want). When training the model in one go (i.e. jointly), we would have no control over how the model handles the utility associated with the semantic attributes. So, to train the Semantic CV-DCM, we first train the semantic regularisation layer. Specifically, we conduct the following steps:

1. We train $\varphi(\circ)$ and $h(\circ)$ while setting $\kappa$ to 1.
2. We freeze the weights of $\varphi(\circ)$ and $h^{sem}(\circ)$, and train the utility parameters associated with the semantic attributes (i.e. $\beta^{sem}$), while setting $\kappa$ to 0.
3. We freeze $\beta^{sem}$ and train $\beta^{res}$, which – in the ideal case – captures the residual utility of attributes and visual cues encoded in images but not part of the semantic layer. Through this sequential training process, we aim to minimise the impact of isomorphic data sharing.

The model is implemented in PyTorch (Paszke et al., 2019). In line with Van Cranenburgh and Garrido-Valenzuela (2023), we used an SGD optimiser with a batch size of 10, an L2 regularisation of 0.1 and a learning rate of 5e-5 for training the model.

Figure 6 shows the results related to training the semantic layer of the Semantic CV-DCM. It depicts the predicted versus the 'true' values (i.e. the values predicted by the GroundingDINO-SAM pipeline) for each semantic attribute for all unique images (2,891) in the test set. In each plot, a linear line is regressed, highlighting the correlation. Additionally, the correlation coefficient is reported in the legend. As indicated by the overall high correlations ($0.37<\rho<0.91$), the model generally does a good job of predicting the semantic attributes. The proportions of buildings and trees are detected with the highest accuracy, while the proportions of water and fences appear more difficult to predict. Even though the predictions are not spot-on, they are sufficiently accurate for the purpose of our paper, which is to examine the factors that shape the distribution of utility derived from street-level conditions. Because the prediction of the semantic attributes using the semantic CV-DCM takes only a fraction of the computational effort as compared to the GroudingDINO-SAM pipeline (which would be infeasible for 1m images), it enables analysing the street-level condition at a city-level scale.

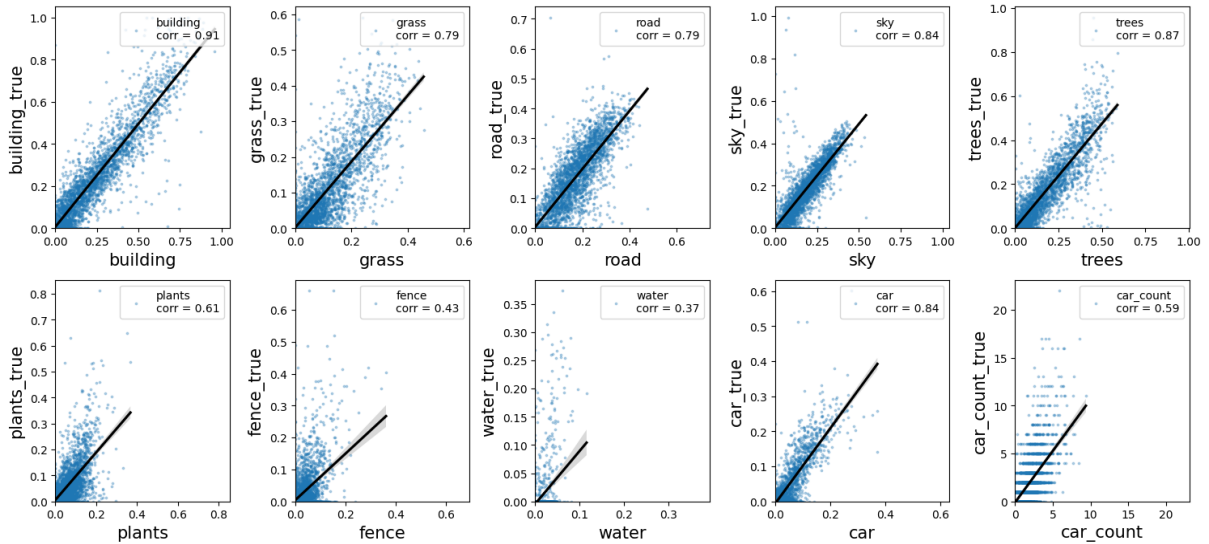

*Figure 6: Results of semantic regularisation*

Table 2 shows the training results related to the choice data. The first column reports the benchmark model, i.e. the plain-vanilla CV-DCM of Van Cranenburgh and Garrido-Valenzuela (2023); the second column shows the results of the Semantic CV-DCM. Looking at the model fit in the test data set, we see that the Semantic CV-DCM fits the data almost equally well as the benchmark model. The fact that the benchmark CV-DCM outperforms the Semantic CV-DCM is expected because the CV-DCM imposes less structure on how the utility is derived from attributes encoded in the image. For instance, the benchmark CV-DCM may learn a nonlinear relation between the number of cars and utility. In contrast, the Semantic DCM imposes a linear relationship. The comparatively better fit on the training dataset suggests that the model slightly overfitsing to the training data. Training the Semantic CV-DCM takes about three times longer than training the benchmark model. This is because the sequential training of the Semantic CV-DCM is conducted in three steps, while the benchmark model can be trained in one go.

Next, we look at the interpretable parameters of the Semantic CV-DCM. Firstly, in line with expectations, we see that the estimates associated with the numeric attributes – housing cost and commute travel time – are virtually the same as those of the benchmark model. Secondly, the signs of the semantic parameters have the intuitively expected signs. In this application, we fixed $\beta_{building}^{sem}$ to zero for normalisation (see section 2.2.2). Given this normalisation choice, we expect positive signs for the parameters associated with e.g. trees, grass, and sky. After all, a positive parameter for a semantic attribute means that increasing its proportion – at the expense of the proportion of buildings (which is the reference category) – will increase the utility derived from the street-level conditions. Likewise, we expect negative signs for the parameters associated with, e.g. roads and fences, because increasing their proportion – at the expense of buildings – can be expected to decrease the utility derived from the street-level conditions.

Since the semantic parameters represent marginal utilities, we can internally compare them to interpret their implications further. For example, the marginal utility of cars is derived from both their count (-0.25) and their proportion in the view (-0.59). This suggests that a single nearby car occupying 50% of the street view is perceived as more detrimental than two cars at a distance occupying only 5% of the view.[2] Likewise, the model suggests that if a building is demolished and three new parking spaces are created on the cleared land, all else being equal, this would negatively impact the street-level conditions.[3] However, if one parking space (with a parked car) and a grassy area of the same size are created instead, the intervention would be utility-neutral.[4] In conclusion, the model offers insights into the importance of various semantic attributes in shaping street-level conditions. That being said, gaining a deeper understanding requires examining the images alongside the model's predictions. This is done through our case study, which is presented in section 3.

---

[2] Because $(\beta_{car\_p} \cdot 0.50 + \beta_{car\_c} \cdot 1) > (\beta_{car\_p} \cdot 0.05 + \beta_{car\_c} \cdot 2)$
[3] Because $\beta_{car\_p} < \beta_{building}$
[4] Because $\beta_{building} \cdot x \approx \left(\beta_{car\_c} \cdot 1 + \beta_{car_p} \cdot \frac{1}{2}x + \beta_{grass} \cdot \frac{1}{2}x\right)$

*Table 2: Training results semantic CV-DCM*

| Data set | | | CV-DCM (benchmark) | Semantic CV-DCM |
|---|---|---|---|---|
| Train data $N = 9{,}784$ | | Parameters | 86m | 86m |
| | | Log-likelihood | -5,724 | -5,572 |
| | | $\rho^2$ | 0.156 | 0.17 |
| | | Cross entropy | 0.585 | 0.570 |
| | | Computation time | ~1.5 hr | ~6.0 hr |
| Test data $N = 1{,}948$ | | Log-likelihood | -1137.6 | -1145.4 |
| | | $\rho^2$ | 0.158 | 0.152 |
| | | Cross entropy | 0.585 | 0.588 |
| Interpretable model parameters | | | Est | Est |
| | Numeric Attr. | $\beta_{hhcost}$ | -0.94 | -0.94 |
| | | $\beta_{tt}$ | -0.23 | -0.24 |
| | Semantic Attr. | $\beta_{car\_c}$ (count) | | -0.25 |
| | | $\beta_{car\_p}$ (proportion) | | -0.59 |
| | | $\beta_{building}$ (proportion) | | 0.00* |
| | | $\beta_{grass}$ (proportion) | | 0.96 |
| | | $\beta_{road}$ (proportion) | | -0.59 |
| | | $\beta_{sky}$ (proportion) | | 1.42 |
| | | $\beta_{trees}$ (proportion) | | 1.40 |
| | | $\beta_{plants}$ (proportion) | | 1.05 |
| | | $B_{fence}$ (proportion) | | -0.81 |
| | | $B_{water}$ (proportion) | | 0.13 |
| | | $B_{unsegmented}$ (proportion) | | -0.25 |

*\* Fixed to zero for normalisation*

## 2.3. Model application and aggregation

In the application, we utilise the trained Semantic CV-DCM model to compute utilities derived from the street-level conditions based on the street-level images of Rotterdam. Notably, in the application, we exclude utilities associated with the numeric attributes 'commute travel time' and 'housing cost', which were part of the training data but are unimportant to our application, in which we want to examine the spatial distribution of utility derived from street-level conditions. Excluding the numeric attributes thus allows us to isolate the utility levels that reflect the street-level conditions.

Step 3 of the methodology (see Figure 1) involves the spatial aggregation of the results. In our application, we spatially aggregated at the postcode 5 level, henceforth called PC5. In the Netherlands, the postcodes consist of four digits followed by two letters (e.g., 1234 AB). The PC5 level (e.g. 1234A) typically corresponds to a small neighbourhood, with an average area equivalent to approximately 335 square metres. This level of aggregation provides a balance between having a sufficient number of images per spatial unit and maintaining a high resolution in the spatial map. The area of study in Rotterdam comprises 867 PC5 areas, with a median of 152 images per area.

## 3. Results

This section comprises three parts. First, Section 3.1 examines the joint distributions of the semantic attributes across space. Understanding their multivariate nature helps to interpret the spatial patterns of street-level conditions better. Then, Section 3.2 presents the main result: the spatial distribution of the utility derived from residential street-level conditions in Rotterdam. Finally, Section 3.3 investigates the contribution of the semantic attributes to its predictions for specific areas by leveraging the explicit nature of the Semantic CV-DCM.

### 3.1. Joint distributions of semantic attributes

Figure 7 shows the bivariate relationships between the semantic attributes as well as the total utility derived from the street-level conditions aggregated at the PC5 level. Each scatter plot shows the pairwise relationships between these attributes, with histograms on the diagonal representing their individual distributions.

Figure 8 provides the following insights. Firstly, the observed correlations between semantic attributes are intuitive. For instance, grass and trees are positively correlated, while buildings and grass are negatively correlated. Also, in line with expectations, we see that the number of cars (car_count) and the proportion of cars (car) correlate positively. These expected correlations further support the notion that the Semantic CV-DCM generally does a good job of predicting the semantic attributes and generalises well towards the images of Rotterdam. Secondly, the correlations between the attributes are not excessively high. This is important for ensuring that the individual effects of these attributes can be disentangled in the model without significant multicollinearity issues. Thirdly, histograms on the diagonal reveal that the semantic attributes are generally normally distributed. Hence, none of the semantic attributes are bimodal or exhibit significant outliers. Also, the distribution of the utility derived from street-level conditions (shown in the bottom-right corner) follows the same pattern. Utility values range from -0.43 to 1.11, with a mean of 0.28. To interpret these numbers, it is important to note that utility does not have an absolute scale. In RUM-based discrete choice models, only differences in utility are meaningful (Train, 2003). In other words, we cannot interpret the sign of the utility level itself, only its relative position within the distribution. For example, a utility level above 0.28 indicates that the street-level conditions in that area are more attractive than average.

Finally, one correlation that warrants further discussion is that of the attribute 'building'. The strong negative correlation between building and utility ($\rho$ = -0.87) may initially seem counterintuitive, given that we set $\beta_{building}^{sem} = 0$ for normalisation. However, this result becomes more intuitive when considering that an increase in the proportion of buildings necessarily leads to a decrease in the proportion of other semantic attributes. Figure 7 illustrates that grass, trees, and sky, in particular, correlate negatively with the proportion of buildings. Moreover, the marginal utilities associated with grass, trees, and sky are positive (see Table 3). So, increasing the proportion of buildings usually leads to a decrease in the proportion of grass, trees, and sky, which means a negative effect on the overall utility level.

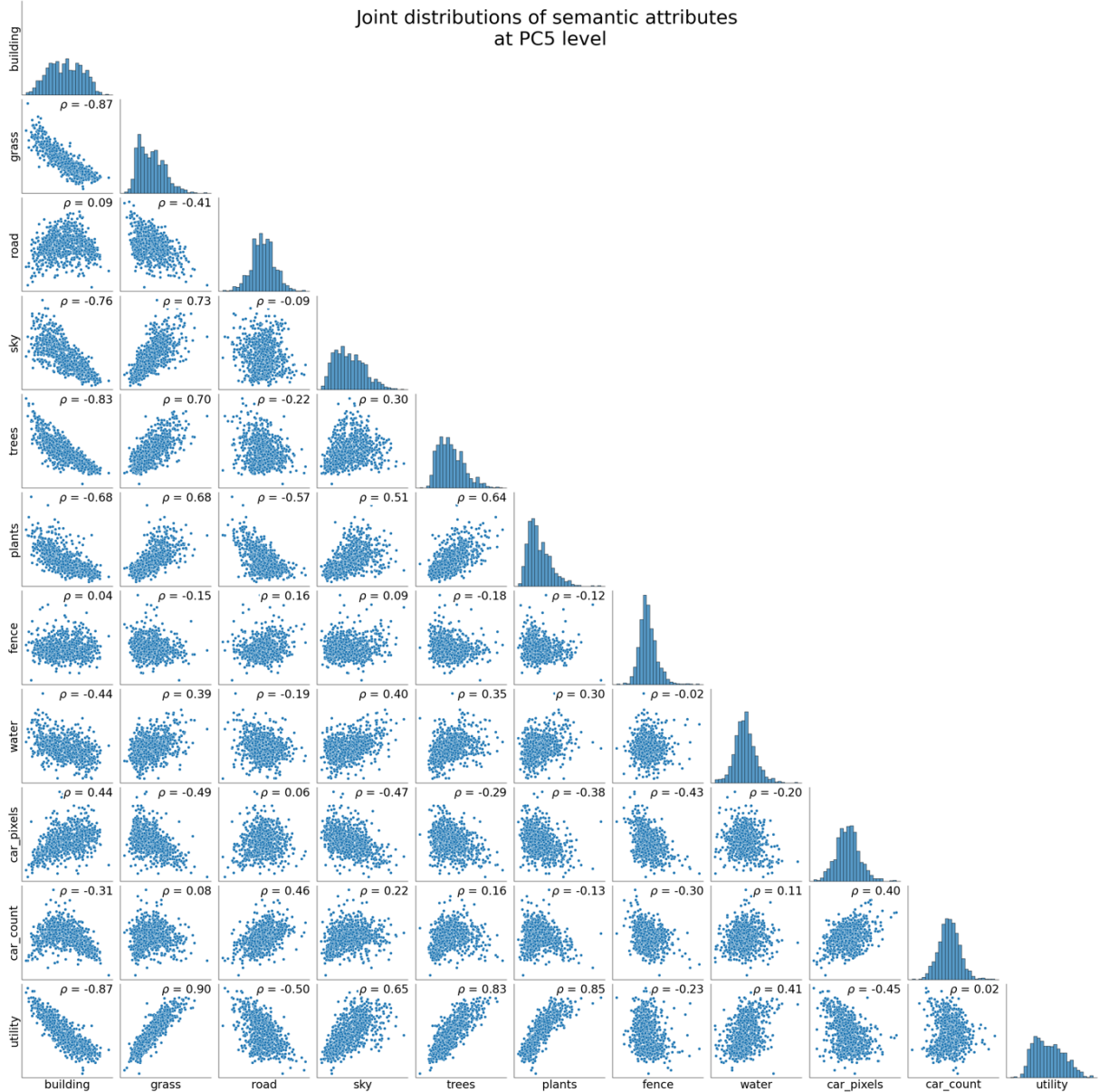
*Figure 8: Joint distribution of semantic attributes*

### 3.2. Spatial distribution of residential street-level condition

The middle plot of Figure 7 shows the main result of this study: the spatial distribution of the utility derived from street-level conditions to residential location choices in Rotterdam. The colour scale is such that the colour red indicates a low utility derived from the residential street-level conditions, and the colour green indicates a high utility derived from the residential street-level conditions. Figure 7 reveals several key insights into the distribution of the street-level conditions in Rotterdam.

Firstly, the city centre (encircled in cyan), which comprises the neighbourhoods 'CS-Kwartier', 'Stadsdriehoek', 'Cool', 'Oude Westen', 'Dijkzigt', 'Nieuwe Werk', exhibits comparatively poor street-level conditions. Except for 'Dijkzigt' and 'Nieuwe Werk', they have below-average street-level conditions. This suggests that the high real-estate prices typically associated with central locations cannot be attributed to the attractiveness of street-level conditions. It indicates that factors other than the immediate street environment, such as proximity to amenities, employment opportunities, or connectivity, play a more significant role in driving up property values in the city centre. The comparative attractiveness of the residential street-level conditions in 'Dijkzigt' and 'Nieuwe Werk' is explained by, respectively, the presence of the museum district and a medium-sized city park.

Secondly, street-level conditions vary considerably, even within small areas. For example, in the neighbourhood 'Hillegersberg Noord' (at the top of the plot), a single dark red spot stands out in an otherwise attractive area, illustrating how conditions can rapidly improve or deteriorate even within close proximity. So, even though 'Hillegersberg Noord' is generally considered an upscale and attractive neighbourhood, this area contains pockets where the street-level conditions are not appealing.

Thirdly, the best residential street-level conditions are found on the city's edges, located particularly near parks and green areas. This spatial pattern reflects the city's spatial hierarchy, where the older, high-density areas dominate the core, which is surrounded by more suburban or commercial/industrial zones. Notable examples include areas near 'Kralingse Bos' in the North and 'Charlois Zuidrand' in the South, as well as residential neighbourhoods with abundant greenery, including trees and plants, such as 's-Gravenland' in the East. These findings underline the importance of greenery in contributing to the attractiveness and utility derived from street-level conditions for the residential location choice.

Fourthly, and perhaps most surprisingly, the southern neighbourhoods of Rotterdam perform moderately well in terms of residential street-level conditions. In common parlance, the southern neighbourhoods, such as 'Bloemhof', 'Tarwewijk', and 'Pendrecht', are perceived as 'probleemwijken' (problem areas) due to higher poverty and crime rates. However, these results show that despite the challenges in these neighbourhoods, the street-level conditions there are not as poor as one might expect. In other words, there are positive aspects to the street-level conditions in these neighbourhoods that may not be immediately apparent in conventional socioeconomic analyses.

### 3.3. Areas of interest

Because the Semantic CV-DCM treats the utility of the semantic attributes in an explicit way, it allows us to investigate the contribution of factors for its predictions for specific areas. In this subsection, we focus on the six blue-encircled PC5 areas shown in Figure 7. These areas are selected because they feature the highest proportion of six relevant semantic attributes and, therefore, make interesting cases. To provide the reader with a better grasp of these areas, collages with random samples of street-level images taken from these areas can be found in Appendix A. Additionally, Appendix B shows the locations of the semantic attributes for these areas within the distribution of the city as a whole.

The bar plots on both sides of Figure 7 show how the explicit handling of semantic attributes by the Semantic CV-DCM can be leveraged to gain deeper insights into the street-level utilities that it predicts. More specifically, the bar plots show how these areas deviate from the mean utility. These deviations are computed by comparing the mean utility derived from each semantic attribute for the city as a whole with that of the area under investigation. The bar outlined in orange highlights the semantic attribute for which this PC5 area shows the greatest proportion. For instance, PC5 area '3054H', in neighbourhood 'Hillegersberg Noord', has the highest proportion of cars, while PC5 area '3056L' features the highest proportion of water. One exception to this is PC5 area '3062T' (shown in the top right). This area features the highest proportion of buildings. However, since we used the proportion of buildings for normalisation (i.e. we fixed $\beta_{building}$ = 0), this semantic attribute is not displayed in the bar plots (see the last paragraph of section 3.1).

The bar plots provide several insights. Firstly, we see that three out of the six areas under investigation, '3054H', '3062T', and '3078N', achieve a by and large average street-level utility (between 0.03 and 0.26). This is explained by semantic attributes that work in opposite directions. Specifically, we see a mix of red and green bars for '3054H', '3062T', and '3078N'. For example, in '3062T', the absence of trees reduces the street-level utility compared to the mean, but relatively few roads and cars compensate for this. Vice versa, in areas '3054H' and '3078N', the street-level utility is lowered because of the above-average presence of cars, but here, the presence of trees compensates for this. Hence, we see that attributes with positive and negative marginal utilities tend to balance each other out. Furthermore, we see that the areas '3056L' in 'Terbregge' and '3084K' in 'Zuiderpark' both feature highly attractive street-level conditions. Interestingly, the bar plots show that different attributes explain this. PC5 area '3056L' achieves a high utility because of the abundant presence of sky and grass. In contrast, PC5 area '3084K' features an average amount of sky. Instead, its high utility is primarily explained by an abundance of trees. Finally, PC5 area '3023A' features comparatively unattractive street-level conditions. This is explained by a combination of few trees, plants and grass, and many roads.

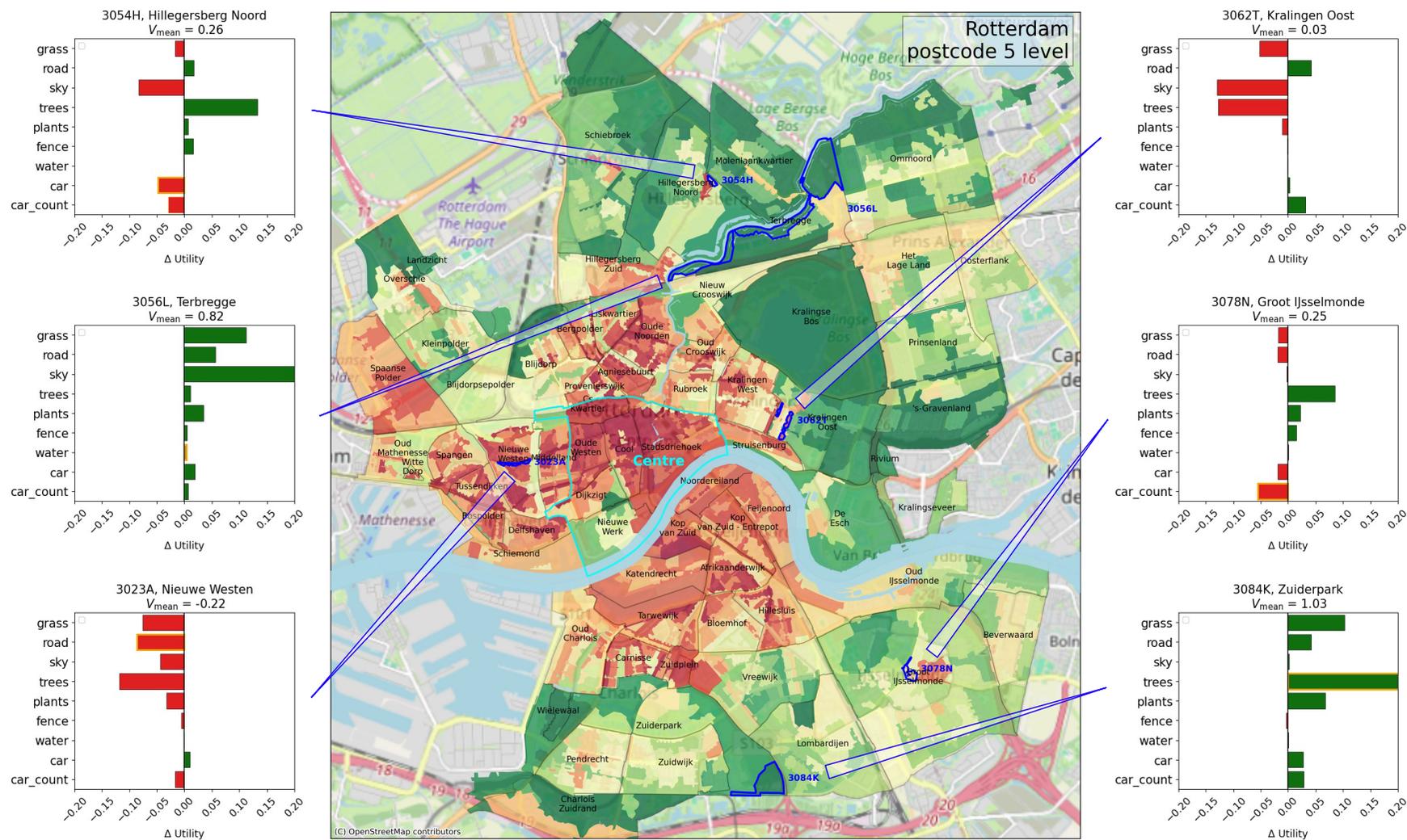

Figure 7: Spatial distribution of residential street-level conditions utility (from images) in Rotterdam (middle). Locations of areas with the highest proportion of semantic attributes are encircled in blue. Impact on street-level utility compared to the mean is shown on both sides for selected areas.

## 4. Conclusion and discussion

This utilises recently proposed computer-vision enriched discrete choice models (CV-DCMs) to provide new insights into the spatial distribution of utility derived from street-level conditions. Thereby, it advances the residential location choice literature by offering insights into the role street-level conditions play in these decisions. Additionally, it complements previous research on urban environment perceptions based on street-level imagery by introducing a preferences-based counterpart. Finally, it makes a methodological contribution by extending the CV-DCM models with a semantic regularisation layer. This layer extracts key semantic attributes from images, which the model then uses to predict the residential location choices. Thereby, we elucidate the computer vision part of the CV-DCM and eliminate the need for separate image processing pipelines.

Several limitations of this study should be acknowledged, providing avenues for future research. Firstly, the utility levels predicted by the Semantic CV-DCM represent mean population preferences. However, residential location preferences are inherently heterogeneous (Smith & Olaru, 2013). Adding to this complexity, individuals may self-select residential locations that align with their specific preferences, such as street-level conditions (Cao, 2014; Van Wee, 2009). Future studies could further investigate the impact of heterogeneous preferences and self-selection effects, potentially through the use of microsimulation (Waddell, 2002). This would enable the creation of spatial maps that reflect how individuals derive utility from the street-level conditions in their living environments.

Secondly, the CV-DCM model used in this research is trained on images captured from a perpendicular angle to the street. This orientation may miss certain street-level attributes that become apparent from a wider or more comprehensive view. For instance, a line of trees might make a perpendicular view feel 'closed,' while a parallel or side-angle perspective could convey a sense of openness. Additionally, narrow perpendicular views may obscure gestalt features, such as the sense of common fate created by a row of trees. Future research could consider using wide-angle street-level images or full panoramic views to capture a more comprehensive view of the street-level conditions.

Thirdly, and related to the previous point, while this study focuses on the utility derived from street-level conditions in residential location choice analyses, it does not account for the potential utility derived from the visual appearance of the broader neighbourhood. The utility people derive from street-level conditions may interact with their perceptions of the wider area. For example, in the case of 'Hillegersberg Noord', a neighbourhood featuring a small area with unattractive street-level conditions, residents might still derive higher-than-expected utility from these conditions due to the overall attractiveness of the larger area, as people can be biased to see what they want to see (Balcetis & Dunning, 2006).

Finally, the images used in this study are taken at ground level, meaning that our model predicts utility derived from street-level conditions at this specific perspective. However, in areas with high-rise buildings, residents may also derive utility from views of the skyline or the openness provided by higher-altitude perspectives. The current study does not consider these factors, but future research could explore the utility residents derive from such views.

**Acknowledgement**
This work is supported by the TU Delft AI Labs programme. Furthermore, the authors express gratitude to Lanlan Yan for her contributions to the early development of the paper. The authors also would like to thank civil servants at the municipality of Rotterdam, particularly Tommie Perenboom, for their assistance in interpreting the substantive results of the case study.
**Declaration of generative AI and AI-assisted technologies in the writing process**
During the preparation of this work the author(s) used ChatGPT-4o in order to improve the readability and language of the manuscript. After using this tool/service, the author(s) reviewed and edited the content as needed and take(s) full responsibility for the content of the published article.

# Appendix A: Visual snapshots of areas of interest

**PC5 area '3054H' Hillegersberg Noord (highest for attribute 'car')**

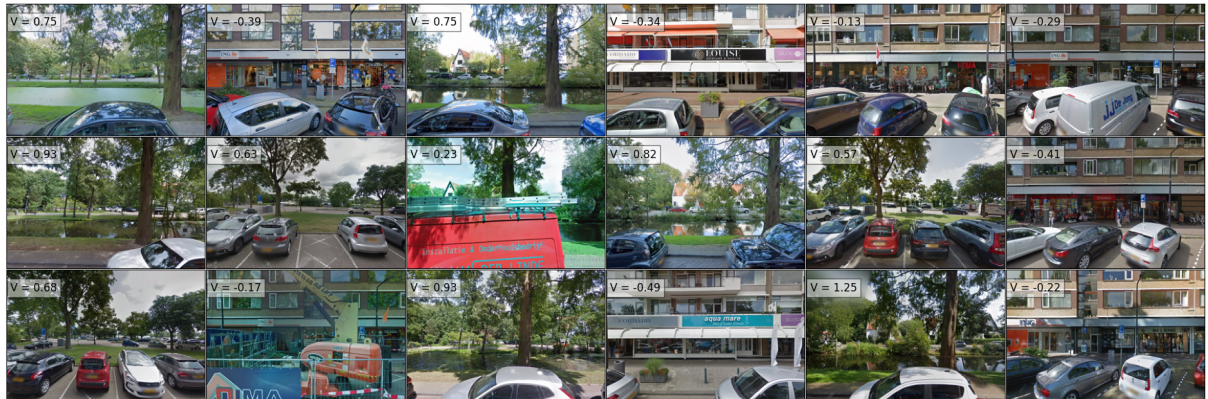

(image source: Google)

**PC5 area '3056L' Terbregge (highest for attribute 'water')**

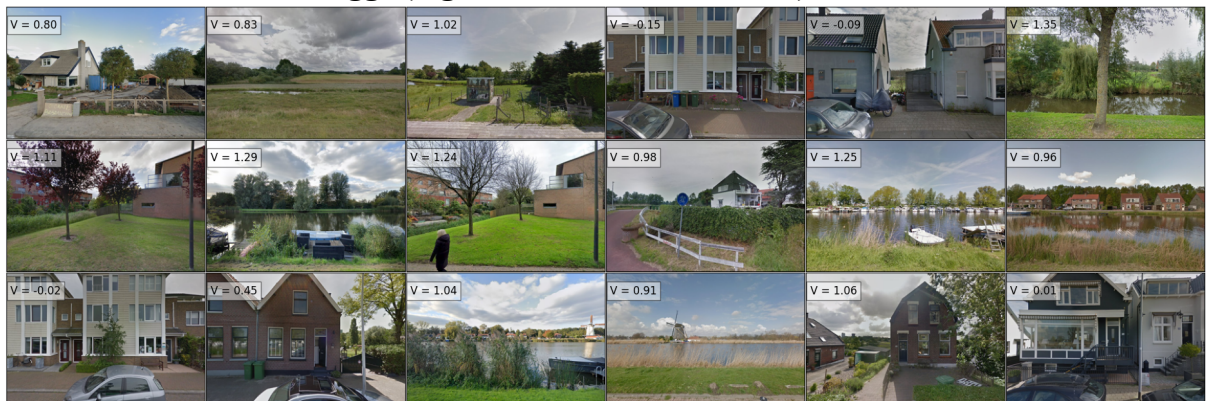

(image source: Google)

**PC5 area '3023A' Nieuwe Westen (highest for attribute 'road')**

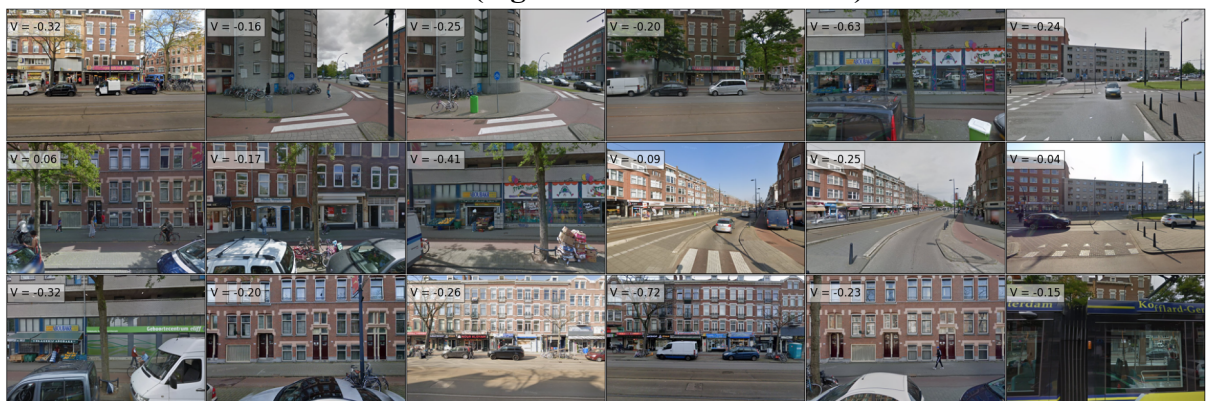

(image source: Google)

**PC5 area '3062T' Kralingen Oost (highest for attribute 'building')**

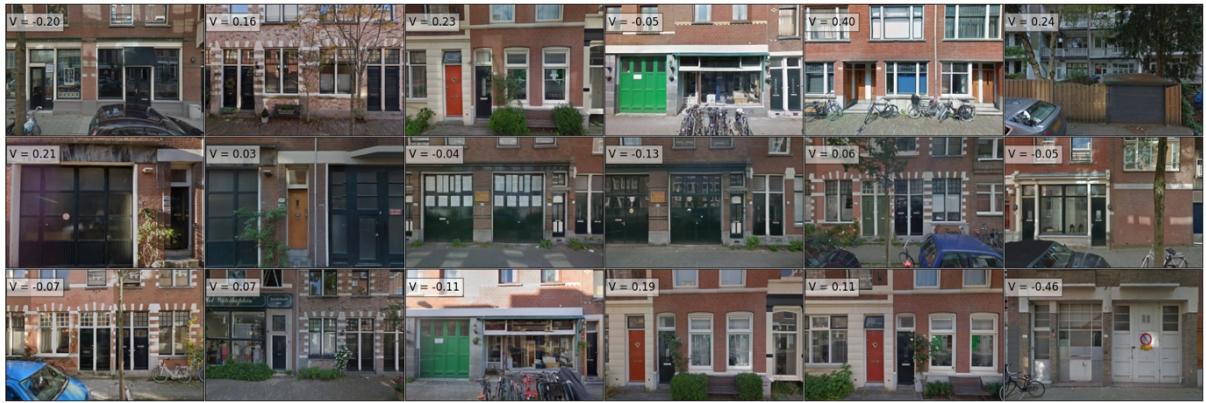

(image source: Google)

**PC5 area '3078N' Groot IJsselmonde (highest for attribute 'car_count')**

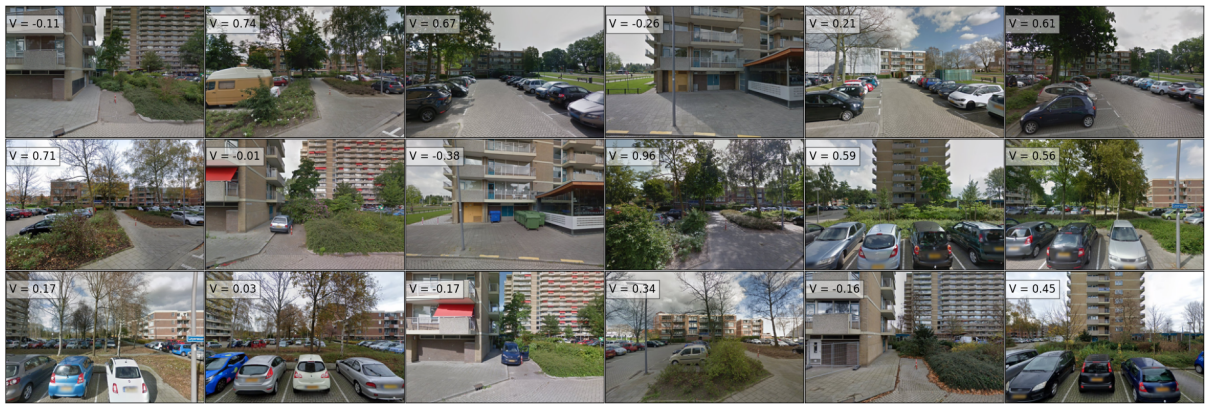

(image source: Google)

**PC5 area '3084K' Zuiderpark (highest for attribute 'trees')**

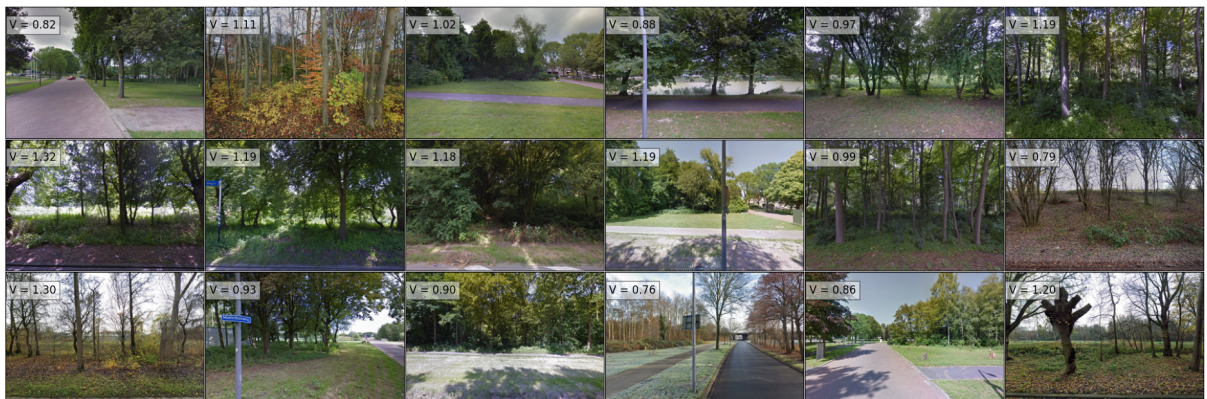

(image source: Google)

# Appendix B: Location of semantic attributes within the overall distribution

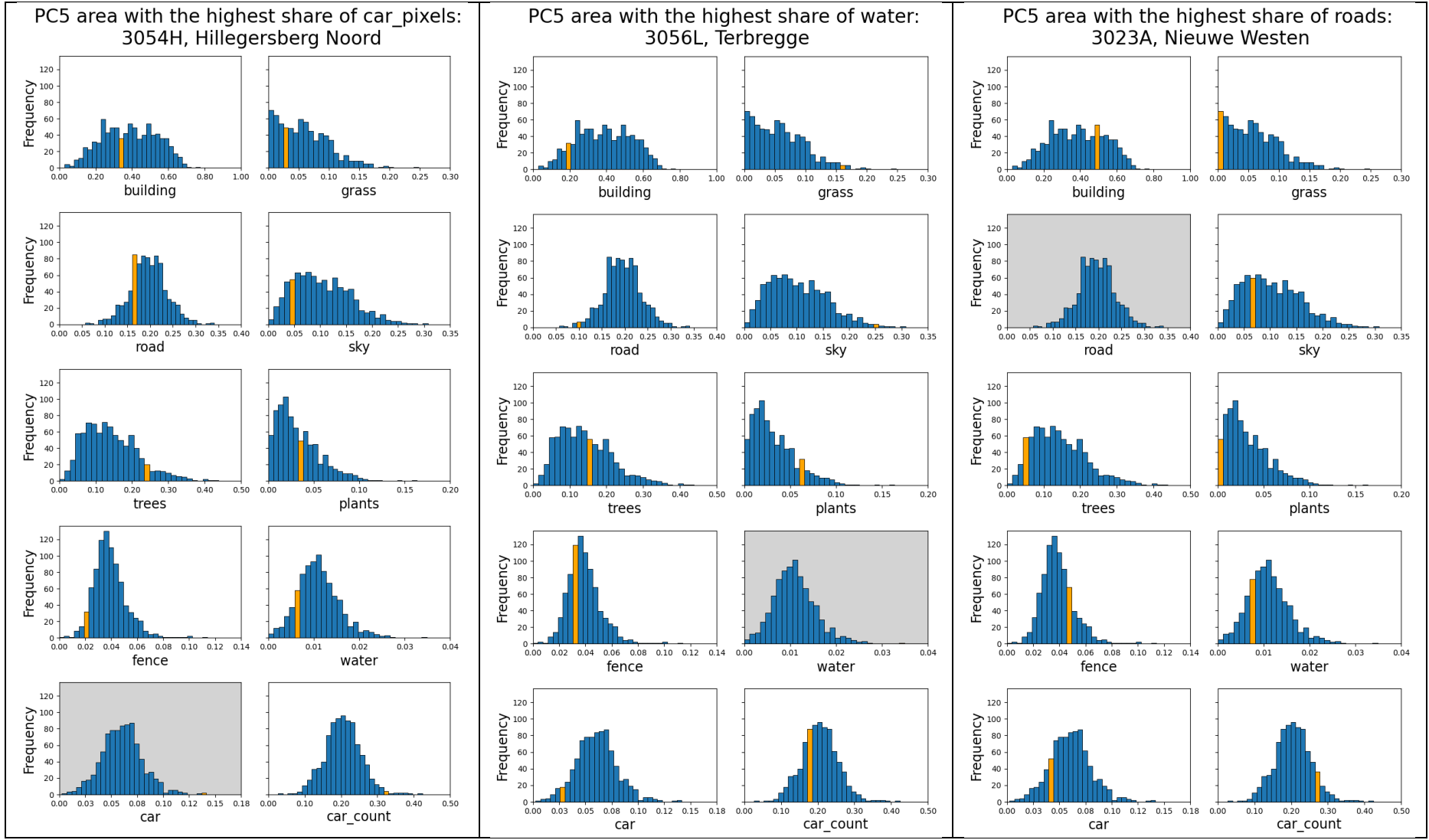

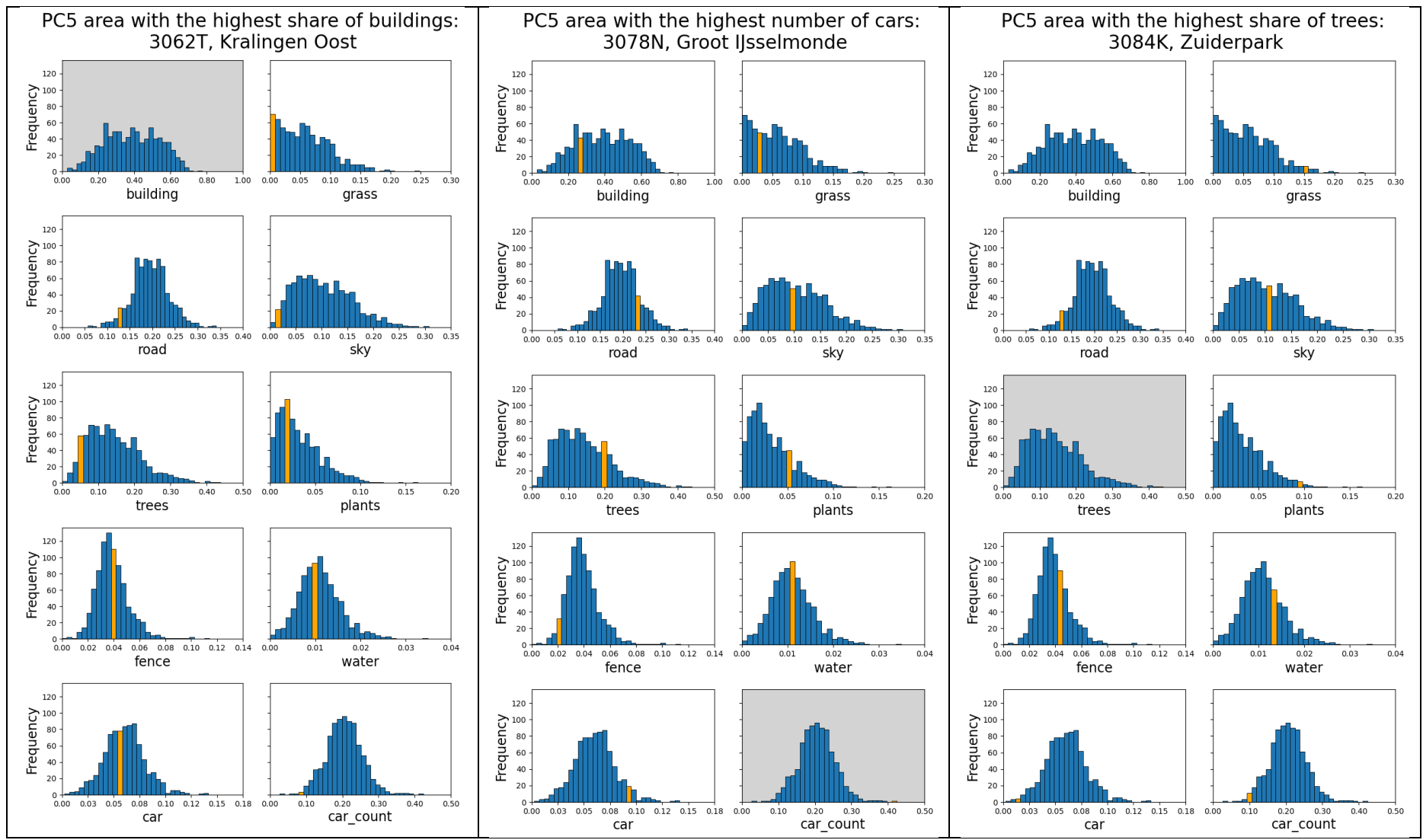